\documentclass[10pt,twocolumn,letterpaper]{article}

\usepackage{cvpr}              
\definecolor{cvprblue}{rgb}{0.21,0.49,0.74}
\usepackage[pagebackref,breaklinks,colorlinks,allcolors=cvprblue]{hyperref}
\usepackage{algorithm}
\usepackage{algpseudocode}
\usepackage{makecell}
\usepackage{multirow}
\usepackage{diagbox}
\usepackage{xcolor}

\def\paperID{11901} 
\def\confName{CVPR}
\def\confYear{2026}

\title{Enhancing Accuracy of Uncertainty Estimation in Appearance-based Gaze Tracking with Probabilistic Evaluation and Calibration}

\author{
Qiaojie Zheng$^{1}$ \quad
Jiucai Zhang$^{2}$ \quad
Amy Zhang$^{1}$ \quad
Xiaoli Zhang$^{1}$\\
$^{1}$Colorado School of Mines, USA\\
$^{2}$GAC R\&D Center Silicon Valley, USA\\
{\tt \{zheng, xlzhang\}@mines.edu, \{zhang.amy666666, zhangjiucai\}@gmail.com}}
\begin{document}
\maketitle
\begingroup
\renewcommand\thefootnote{}
\footnotetext{Code available at \href{https://github.com/GrantZheng86/Gaze_estimation_uncertainty_calibration}{project page}.}
\endgroup
\begin{abstract}

    Accurate uncertainty estimation is essential for reliable appearance-based gaze tracking. However, domain shifts between training and testing often lead to incorrect uncertainty estimates, which is a problem overlooked in existing uncertainty-aware gaze tracking models. To overcome this problem efficiently, we formulate uncertainty estimation as a conditional distribution problem and treat the correction process as an output-level conditional distribution matching task. We therefore introduce a data-efficient post-hoc calibration method to align the predicted, high-error conditional distribution with the empirically observed distribution extracted from a small set of calibration samples. To more faithfully assess the accuracy of the resulting uncertainty estimates, we further introduce a new metric, Coverage Probability Error (CPE), to quantify the distribution-level mismatch between prediction and observation. We validate the calibration procedure across four domain shift scenarios to demonstrate improved uncertainty accuracy and its practical benefits.

\end{abstract}

\section{Introduction} \label{intro}
Appearance-based gaze tracking is a task characterized by significant uncertainty~\cite{Zhong2024}. Factors such as subject appearance, lighting conditions, and camera parameters can significantly affect model performance~\cite{Zhang2015}. For reliable mission-critical applications, such as driver monitoring systems, mere point estimates of the gaze angle are insufficient; the accompanying inference uncertainties must be provided to assess gaze-tracking reliability.

Equipping models with uncertainty awareness is challenging since the ground truth of uncertainty is not available for each training sample. Common approaches estimate uncertainties through probabilistic modeling~\cite{Her2023, Chen2019} and quantile regression~\cite{Kellnhofer2019}, to obtain predictive variances or prediction range to represent uncertainty. These methods model gaze-tracking uncertainty as input-conditioned variance or prediction range, allowing uncertainty estimation without explicit uncertainty labels. 

\begin{figure}[t]
    \centering
    \includegraphics[width=0.88\linewidth]{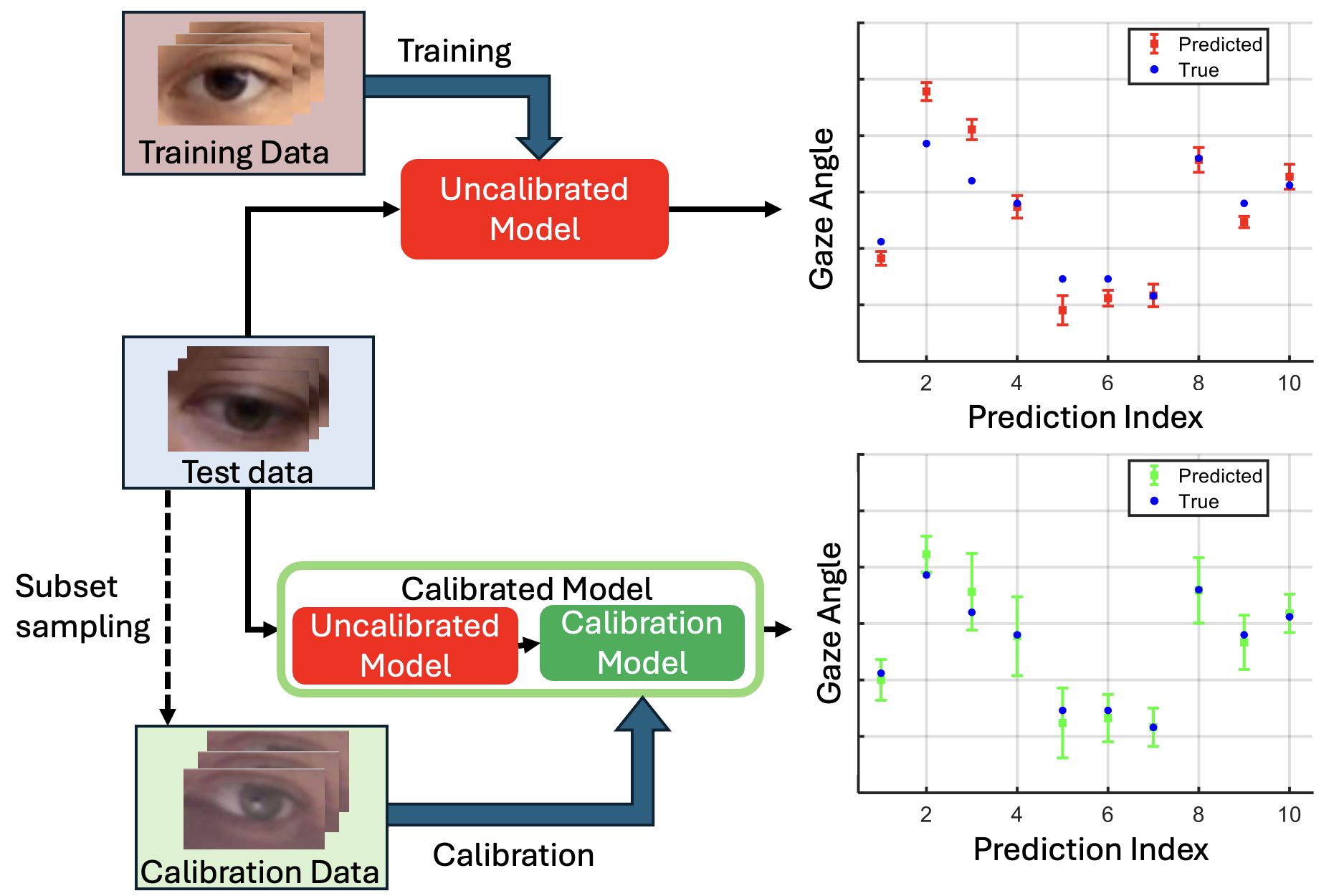}
    \caption{Inaccurate uncertainty estimation of existing uncertainty-aware gaze tracking models (top) and our calibration procedure (bottom). Uncertainty estimates from existing models provide inaccurate probabilistic predictions and often fail to capture the ground truth gaze angle values (blue dot in angle prediction plots) due to domain shifts. We introduce a calibration procedure to compensate for such shift. The calibrated model produces accurate uncertainty estimates, capturing most ground-truth gaze angles for improved downstream applications.}
    \label{fig:Figure_1}
\end{figure}

Despite their uncertainty estimation capabilities, existing uncertainty-aware approaches are vulnerable to domain shifts. Since their uncertainty models are derived solely from the conditional distribution of the training data without explicit regularization, these models inherently encode data-specific bias. Consequently, the predicted uncertainty magnitudes are inaccurate. The numerical uncertainty values cannot be used directly, as the learned estimation rule between image inputs and uncertainty scale is domain dependent. As a result, the outputs are only reliable for relative ranking within domain. This limitation is evident in nearly all uncertainty related works in gaze tracking~\cite{Her2023, Kellnhofer2019, Wang2024-zx, Cai2023-xi, Zhong2024}, as their applications and evaluations are purely on domain-specific ranks, rather than on numerical uncertainty values.

Although parameter-level adaptation methods such as meta-learning and transfer learning could, in theory, correct uncertainty predictions, they face substantial data demands. Correcting uncertainty requires large amounts of target-domain data to relearn the conditional predictive distribution, far more than the simple mapping adjustments typical in standard adaptation scenarios~\cite{Feinstein2021-wj, Abdar2021-cd}. To reduce this burden and enable efficient correction, we avoid modifying model parameters and instead apply post-hoc output calibration so the calibrated distribution aligns with the observed one. The calibration procedure first quantifies the discrepancy between predicted and observed distributions, then derives a correction rule that reduces this gap so the predicted distribution matches the empirical one. During testing, this rule is applied directly to the outputs of the uncalibrated, training-domain-specific model to produce reliable uncertainty estimates under domain shifts. The overall calibration process is illustrated in Figure~\ref{fig:Figure_1}.

\begin{figure}[t]
    \centering
    \includegraphics[width=0.65\linewidth]{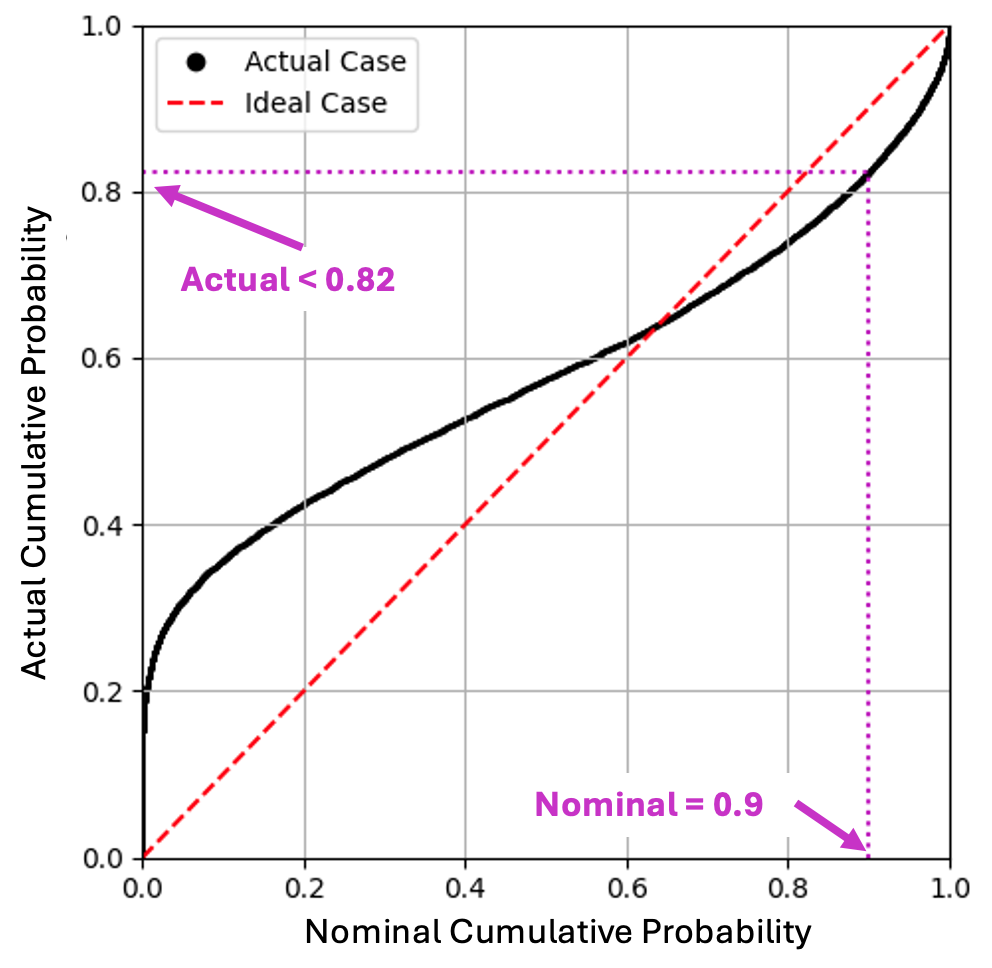}
    \caption{Illustration of uncertainty errors. Areas between the actual and ideal case denote the severity of errors. An error example is shown at 0.9 nominal cumulative probability. The actual cumulative probability is less than 0.82, leading to an 8\% error. Root mean squared of the areas formed between the two lines is CPE. }
    \label{fig:Figure_2}
\end{figure}

To more faithfully reflect the calibration effect and assess the quality of uncertainty estimates, we introduce a new evaluation metric, Coverage Probability Error (CPE), to evaluate the difference between predicted and observed distribution. An ideal uncertainty model produces distributions that align perfectly with the observed ones, any discrepancies, as illustrated by the quantile value coverage difference shown in Figure~\ref{fig:Figure_2}, reflect inaccuracies in uncertainty models. CPE leverages such difference to evaluate the model performance. We use CPE instead of the commonly used error–uncertainty correlation (EUC) metric~\cite{Kellnhofer2019, Her2023} because EUC does not reliably reflect uncertainty accuracy. As uncertainty only stems from aleatoric and epistemic reasons~\cite{kendall2017uncertainties}, instead of error-related ones, the correlation between angular error and uncertainties is spurious. Hence, EUC provides only an observed association between angular error and uncertainty and cannot be used to draw conclusions about the quality of the uncertainty model~\cite{Rasmussen2023-zt, Stahl2020-ct}, whereas CPE directly evaluates how well the predicted distribution matches the observed distribution for proper performance assessment. 




We evaluate our calibration method under four domain shift scenarios, spanning cross-subject and cross-dataset settings using two widely studied gaze-tracking datasets, MPIIGaze and RTGene. These two datasets exhibit distinct but internally consistent image quality characteristics, providing diverse and well-defined domain shifts for thorough testing of the calibration process. Experimental results show that our calibration procedure consistently corrects inaccurate uncertainties and yields substantially more reliable uncertainty estimates across all scenarios. We also observe that the calibrated models yield improved angular accuracy when using the calibrated 50\% quantile values (i.e. median) to represent the gaze angle, as the median provides a more robust gaze angle prediction. Finally, we showcase the practical benefit of the calibration approach through a case study that estimates 95\% confidence intervals (CI).  In short, our contributions are as follows:
\begin{itemize}
    
    \item Introduce calibration to reduce uncertainty estimation inaccuracies caused by domain shifts in gaze tracking.
    \item Introduce a proper evaluation metric for accurate assessment of uncertainty estimates, providing a more faithful measure of the uncertainty-aware model's performance compared to existing error-uncertainty correlation. 
    \item Demonstrate the effectiveness of the correction model using two Convolutional Neural Network (CNN) models and two datasets.  
    \item Present a case study that compares the accuracy of 95\% CI to illustrate the practical utility of calibration.
\end{itemize}

\section{Related Work}
\subsection{Appearance-based Gaze Tracking} 
\label{sec:related_1}

Appearance-based gaze tracking directly predicts gaze angles from eye or face images using deep learning, without explicit feature extraction such as pupil detection or corneal reflections~\cite{Sugano2013}. With the growth of large datasets, such as Columbia Gaze~\cite{CAVE_0324}, MPIIGaze~\cite{Zhang2015, Zhang2017}, and RTgene~\cite{Fischer2018}, and increasingly powerful neural networks~\cite{Zhang2017, Chen2019a, Park2019}, appearance-based methods now surpass model-based approaches in angular accuracy and generalizability.

However, domain shifts, such as lighting condition changes, often cause noticeable drops in prediction accuracy. To address this issue, adaptation techniques such as meta-learning have been proposed. For example, a meta-learning framework~\cite{Park2019} improved cross-dataset generalization, achieving a state-of-the-art angular error of $3.14^\circ$ on the MPIIGaze dataset. Although some recent models, such as L2CS-Net~\cite{10372944}, achieve high accuracy without requiring additional subject-specific training data, large interpersonal appearance variations still limit their performance to around $3.9^\circ$. Despite these advances, most state-of-the-art models lack the capability to provide accurate, real-time uncertainty estimations. This limits their applicability in safety-critical scenarios, where avoiding high-uncertainty predictions for downstream applications is crucial.

\subsection{Uncertainty-Aware Appearance-based Gaze Tracking and Evaluation Metrics}

Models estimate uncertainties through three main approaches, variance-, range-, and contrastive-based approaches. Variance-based methods model uncertainty as the predicted variance of a probabilistic distribution~\cite{Her2023, Chen2019, Zhong2024, Cai2023-xi}, while range-based methods, such as quantile regression, learn to predict intervals directly~\cite{Kellnhofer2019}, and contrastive-based methods infer uncertainty from embedded dispersion~\cite{Wang2024-zx}. Among these three approaches, variance-based approaches provide well-defined probabilistic outputs, making it well-suited for calibration. Range- and contrastive-based methods, however, lack consistent probabilistic meaning. It should be noted that although Monte Carlo dropout~\cite{gal2016dropout} and ensembles are popular approaches in other applications to acquire uncertainties, they are rarely adopted in gaze tracking due to high computation cost. 

Despite their uncertainty-estimation capability, existing gaze-tracking models often produce numerically inaccurate estimates under domain shifts, so prior work relies on relative uncertainty ranking rather than absolute values. For example, Chen et al.~\cite{Chen2019} use uncertainty ranking to identify outliers instead of providing reliable confidence intervals. Moreover, uncertainty-model evaluation in gaze tracking remains unreliable due to flawed metrics. The widely used error–uncertainty correlation (EUC)~\cite{Her2023, Kellnhofer2019} is spurious because it assumes uncertainty is caused by prediction error, rather than by aleatoric or epistemic factors~\cite{kendall2017uncertainties}. As a result, EUC has limited interpretability and fails to faithfully measure uncertainty quality. For instance, Kellnhofer et al.~\cite{Kellnhofer2019} reported an EUC of 0.46 but did not use it to assess uncertainty accuracy, leaving the evaluation inconclusive. Since prediction error and uncertainty are noncausal~\cite{Rasmussen2023-zt}, even a perfect uncertainty model cannot achieve an EUC of 1, underscoring its unreliability.

\section{Methodology}
\subsection{Proper Scoring for Assessing Uncertainty}

A proper scoring metric, by definition, assigns the best score only to a model whose predicted distribution matches the true underlying data distribution~\cite{Gneiting2007}. In uncertainty-aware gaze tracking, the ideal model would produce probabilistic predictions whose distribution perfectly matches the true data distribution. However, comparing distributions is not feasible for individual inferences, as the full distribution is inaccessible at that level.

Instead of assessing the distribution at an instance level, we evaluate the uncertainty accuracy at the population level by analyzing how predicted distributions align with empirical observation across all test samples. An accurate uncertainty-aware model should, for instance, produce 90th-percentile estimation that can contain ground truth gaze angles 90\% of the time. By comparing nominal cumulative probabilities (e.g., 90\%) with the empirical coverage probability, which is defined by the proportion of ground truth values falling below the corresponding predicted quantiles, we assess how well the predicted distribution reflects actual outcomes. This yields a distribution-level measure of the uncertainty quality. Formally, we can express error for each nominal cumulative probability $p$ with equation~\ref{eqn:eqn_1}. 
\begin{equation}
    p_{err}(p)=\left|p-\sum_{t=1}^{T}\frac{I\left\{\theta_t\le F_t^{-1}\left(p\right)\right\}}{T}\right|
    \label{eqn:eqn_1}
\end{equation}

In this equation, $\theta_t$ denotes the ground truth gaze angle value for the $t^{th}$ input sample, $x_t$; $p$ is the nominal probability value in the range of $[0, 1]$, $T$ is the total number of samples involved, $F_t$ denotes the cumulative distribution function (CDF) derived from the probabilistic prediction from the uncertainty-aware model, $H$, on input sample $x_t$, i.e., $F_t = H(x_t)$. $F_t^{-1}$ is the quantile function (inverse CDF) that calculates the quantile value based on the input probability. Here, $I$ denotes the indicator function that equals 1 when the condition is satisfied and 0 otherwise. $\sum_{t=1}^{T}\frac{I\left\{\theta_t\le F_t^{-1}\left(p\right)\right\}}{T}$ represents the probability that the ground truth angle values fall within the estimated boundary, calculated with the quantile function on the nominal probability $p$. We refer to this term as the empirical probability derived from observation. For simplicity, we will denote this term as $\hat{P}\left(p\right) = \sum_{t=1}^{T}\frac{I\left\{\theta_t\le F_t^{-1}\left(p\right)\right\}}{T}$.

Equation \ref{eqn:eqn_1} only defines the error at a single nominal probability. For evaluation of the overall distribution discrepancy, this difference needs to be evaluated over the entire probability range of $[0, 1]$. Inspired by the Brier score \cite{BRIER1950}, we introduce a measure called Coverage Probability Error (CPE), shown in Equation \ref{eqn:eqn_2}, to reflect the root mean square error (RMSE) of coverage discrepancies evaluated across the entire probability range of $[0, 1]$. $n$ in this equation is the number of nominal probabilities at which the CPE will be evaluated. $\frac{i}{n}$ represents the nominal probability, i.e., $p$ in Equation~\ref{eqn:eqn_1}, at which uncertainty errors are evaluated. A larger $n$ enables more precise evaluation of CPE across nominal probability levels, while a smaller $n$ reduces computation time. In this study, we set $n=11$ to balance accuracy and efficiency, evaluating coverage probability at intervals of 0.1 from 0 to 1. The user may choose different $n$ values or introduce weight to put emphasis on different intervals depending on the application. CPE is also relatively interpretable—its magnitude directly reflects how much empirical coverage deviates from nominal confidence levels; for example, a CPE of 0.05 indicates that a nominal 80\% confidence interval effectively covers about 70–90\% of true samples, corresponding to roughly 5\% miscoverage on each side of the interval. Algorithm~\ref{algo:algo1} provides a pseudo-code implementation of the CPE calculation.

\begin{equation}
    CPE=\sqrt{\frac{1}{n}\sum_{i=0}^{n}{p_{err}\left(\frac{i}{n}\right)^2}\ }
    \label{eqn:eqn_2}
\end{equation}

\begin{algorithm} 
\caption{Calculation of CPE}\label{algo:algo1}
\begin{algorithmic}[1]
\Require Uncertainty-aware model $H$
\Require Eye images $X = [x_1, \dots, x_T]$
\Require Gaze angle label $\Theta = [\theta_1, \dots, \theta_T]$, where $\theta_t$ is the gaze angle label for image $x_t$

\Procedure{CalculateCPE}{$H, X, \Theta$}

    \State Initialize $CPE^2 \gets 0$
    \For{all intended probabilities $p$ }
    \State Initialize inclusion count $\mathrm{inc} \gets 0$
    \For{$t = 1$ to $T$}
        \State Calculate quantile function $F_t \gets H(x_t)$
        \If{$\theta_t \leq F_t^{-1}(p) $} \Comment{Check the gaze angle resides in the quantile}
            \State $\mathrm{inc} \gets \mathrm{inc} + 1$ \Comment{Update inclusion count}
        \EndIf
    \EndFor
    \State Calculate the observed probability $p_{obs} = \frac{\mathrm{inc}}{T}$ 
    \State Calculate the model error $p_{err, p} = |p - p_{obs}|$
    \State $CPE^2 \gets CPE^2 + \frac{p_{err, p}^2}{n}$ \Comment{Update $CPE^2$}
    \EndFor
    \State $CPE = \sqrt{CPE^2}$
\EndProcedure
\end{algorithmic}
\end{algorithm}

To visualize the concept of CPE, we created a plot that compares the nominal probabilities with the observed ones in Figure~\ref{fig:Figure_2}. For a perfect uncertainty model, the nominal cumulative probability should match the observed cumulative probability, forming a diagonal line. However, errors in the uncertainty model cause deviations from the ideal case, creating the black curve. This figure also illustrates a specific error case occurring at the nominal cumulative probability of 0.9. Although we expect the quantile values calculated at the nominal 0.9 cumulative probability to encompass 90\% of all gaze ground truth labels, in reality, fewer than 82\% of the ground truth labels fall below these quantiles. This error leads to overconfident predictions and underestimated uncertainty. According to Equation \ref{eqn:eqn_1}, the error of this uncertainty model calculated at the nominal probability of 0.9 is approximately 8\%. Across the full probability range, the CPE value is computed as the square root of the integral of the squared differences between the observed and ideal lines. Models with high CPE values are inaccurate and are formally referred to as miscalibrated models. Unlike EUC, CPE directly measures calibration quality by quantifying discrepancies between predicted probabilities and observed outcomes across the entire probability range, avoiding reliance on potentially misleading correlations.

\subsection{Uncertainty Model Calibration}
To correct miscalibration in uncertainty models, a practical approach is to adjust the nominal probabilities used during inference. For example, if the nominal 0.9 quantile covers only 80\% of ground-truth values, we can use the nominal 0.9 quantile whenever 80\% coverage is needed. In other words, instead of relying on the nominal 0.8 quantile, we substitute the one that actually yields 80\% coverage. This idea can be formalized by using an auxiliary calibration regressor model to store this knowledge. This regressor $R$ learns a mapping, $R:\left[0, 1\right]\rightarrow[0, 1]$, that maps the nominal cumulative probabilities to the actual ones. The corrected CDF is then expressed as $R\circ F$, allowing the estimations to better approximate the true distribution. Because $R\circ F$ does not modify the original uncertainty-aware model, such calibration procedure is applicable to all models that output probabilistic distributions, not only limited to the CNN-based methods presented in this study. 

The regressor's objective is described in Equation \ref{eqn:eqn_3}, where $p_i$ represents the nominal cumulative probability for the $i^{th}$ sample, calculated from the uncalibrated probability output from the uncertainty-aware model, ordered such that $p_i\ \le\ p_{i+1}$. This ordering preserves the inherent monotonicity of any CDF, even when miscalibrated. Accordingly, the calibration model $R$ is constrained to satisfy $R\left(p_i\right)\le R\left(p_{i+1}\right)$ for any realistic corrections. 

\begin{equation}
\begin{array}{l}
\displaystyle \min \sum_{i=1}^{T} \left\| \hat{P}(p_i) - R(p_i) \right\| \\
\mathrm{s.t. } R(p_i) \le R(p_{i+1}), \quad \forall\, p_i \le p_{i+1}
\end{array}
\label{eqn:eqn_3}
\end{equation}

During usage or test time, the quantile values are determined based on the error-adjusted nominal probability $\tilde{p} = R(p)$ rather than the nominal $p$ value. This ensures that predictions align more closely with true values, as described in Equation \ref{eqn:eqn_4}. The left-hand side of the arrow represents the observed coverage, i.e., ground truth gaze angle labels smaller or equal to the quantile, and the right-hand side represents the nominal probability. Therefore, the calibrated estimates are described as $\hat{P}\left(R(p)\right)$.

\begin{equation}
    \sum_{t=1}^{T}\frac{I\left\{\theta_t\le F_t^{-1}\left(R(p)\right)\right\}}{T}\rightarrow p
    \label{eqn:eqn_4}
\end{equation}

To train the calibration regressor model $R$, we need a dataset $D=\left\{\left(p_i,\ \hat{P}\left(p_i\right)\right)\right\},\ i=1..n$, uncalibrated CDF’s error behavior across the full range $p \in[0, 1]$. However, this is challenging since we cannot directly sample based on $p$ values; instead, we must sample from the input space $x$. In this study, we randomly sampled from the testing dataset to create a subset for training the calibration model. By creating this subset, we assume it adequately represents the overall probability distribution of the data.

After collecting the training data, we construct the calibration model $R$ using isotonic regression, thereby respecting the monotonicity of the CDF. This yields a simple, non-parametric calibration model with minimal overhead. Its flexibility also frees the uncalibrated CDF from parametric assumptions imposed by the uncertainty-aware model, allowing a closer match to the true distribution. Other approaches, such as temperature or linear scaling, assume parametric forms that may not capture the nonlinear miscalibration patterns. Median values are used as point estimates instead of means, which are typical for normal assumptions. Because the gaze vector has two components, calibration is performed separately for each. For clarity, Algorithm~\ref{algo:algo2} provides pseudo-code for the process. Here, $\hat{\tilde{\theta}}_{t,quant}$ is the calibrated quantile for the $t^{th}$ input, $\hat{\cdot}$ denotes model predictions, and $\tilde{\cdot}$ denotes calibrated quantiles.

\begin{algorithm}
\caption{Calculate calibrated quantile} \label{algo:algo2}
\begin{algorithmic}[2]
\Require Define the array of nominal probabilities $P = [p_0, p_1, \dots, p_n]$ 
\Require Calculate observed probability array $P_{obs} = [\hat{P}(p_0), \hat{P}(p_1), \dots, \hat{P}(p_n)]$ 
\Require Uncertainty-aware model $H$
\Require Eye image input $x_t$
\Require The nominal probability $p$ 
\Procedure{CalibQuant}{$P, P_{obs}, H, x_t, p$}
    \State Calibration model $R \gets \mathrm{FitIsotonic}(P, P_{obs})$
    \State Error-accounted probability $\tilde{p} \gets R(p)$ 
    \State Cumulative distribution function $F_t \gets H(x_t)$
    \State Error-accounted quantile value $\hat{\tilde{\theta}}_{t,quant} \gets F_t^{-1}(\tilde{p})$
\EndProcedure
\end{algorithmic}
\end{algorithm}

\begin{figure}
    \centering
    \includegraphics[width=1.0\linewidth]{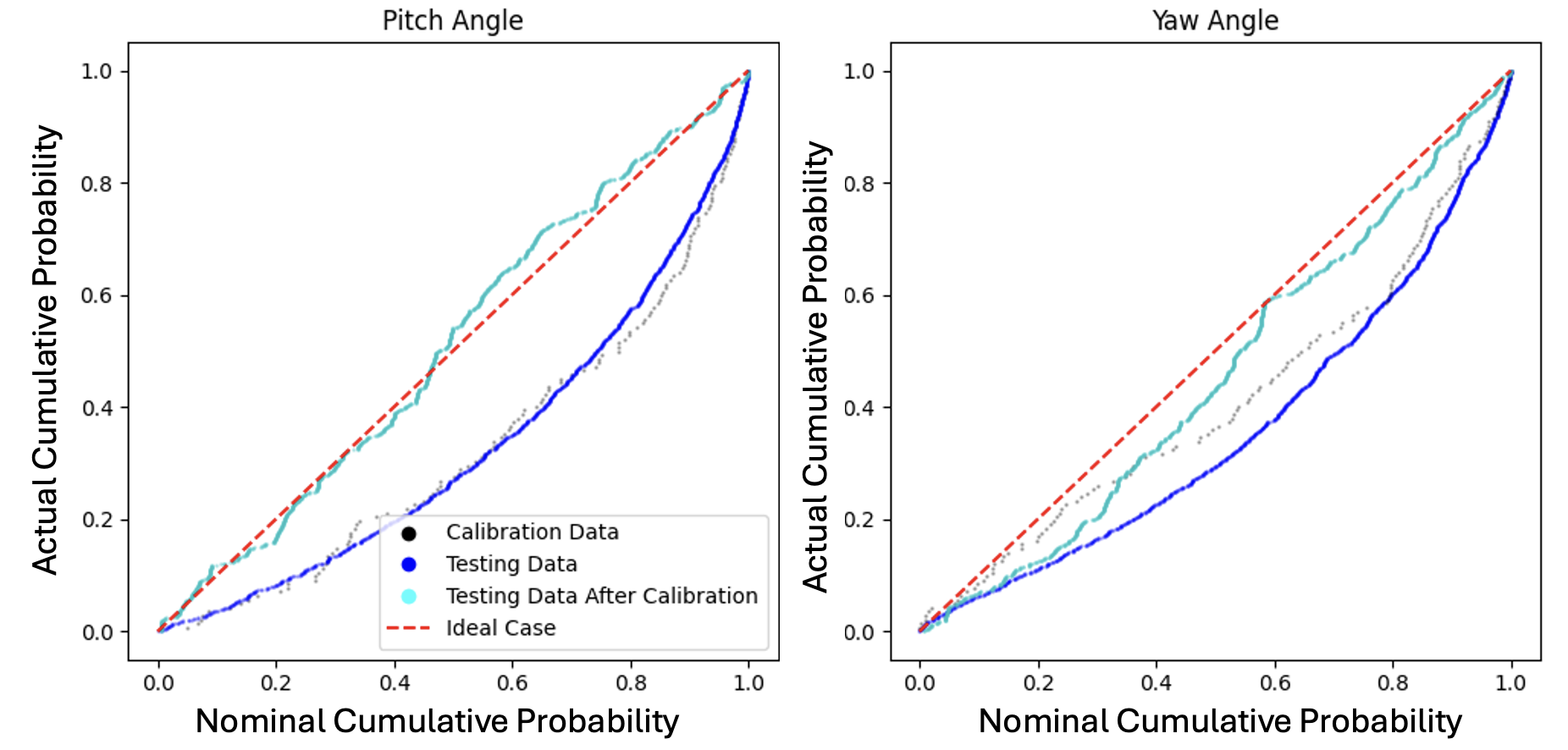}
    \caption{Calibration mechanism overview. Uncalibrated predictions deviate significantly from the ideal due to poor uncertainty estimates. Calibration learns corrections from calibration data with similar error pattern, resulting in predicted distributions that align more closely with the ideal curve}
    \label{fig:calib_mech}
\end{figure}

To help visualize the calibration mechanism, we created a plot comparing the ideal, calibrated, and uncalibrated results in Figure~\ref{fig:calib_mech} based on one of our experiments. The blue and black dots represent images from the same person and are used for testing and calibration, respectively. Since the calibration and testing data are collected from the same person, we assume they follow a similar distribution, so a model calibrated on the calibration data is applicable to the testing data. Before calibration, quantile values from the model’s probabilistic predictions fail to capture the ground truth labels at the intended probability, resulting in large deviations from the ideal, as shown by the blue curve. During calibration, the corrective regressor, $R$, learns to correct errors from the calibration data (black dots) which exhibit patterns similar to those in the testing data. When the calibration model is applied, the erroneous behaviors observed in the testing data caused by model miscalibration can be corrected to produce results that closely follow the ideal case, as shown by the cyan line.

\section{Experiments}
\subsection{Gaze Tracking Datasets}

\begin{figure}
    \centering
    \includegraphics[width=0.7\linewidth]{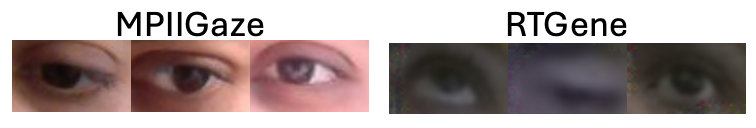}
    \caption{Image samples from the MPIIGaze and RTGene dataset. Images in the RTGene dataset are much noisier than MPIIGaze images, leading to large domain shifts. }
    \label{fig:MPII vs RTGENE}
\end{figure}

Our experiments are designed using dataset combinations that exhibit varying yet internally consistent levels of domain shift, enabling systematic analysis of calibration effectiveness under different shift severities. To this end, we select MPIIGaze~\cite{Zhang2015} and RTGene~\cite{Fischer2018}, two widely used, unconstrained gaze datasets collected under relatively controlled setups, each characterized by distinct but internally coherent imaging conditions. Due to the larger camera–subject distance and additional post-processing, RTGene images are noticeably blurrier than those in MPIIGaze, as illustrated in Figure~\ref{fig:MPII vs RTGENE}. These internally consistent yet externally distinct characteristics make the two datasets well-suited for constructing domain shift scenarios, including cross-subject and cross-dataset ones.

\subsection{Uncertainty-Aware Gaze Tracking Structure}
\begin{figure}
    \centering
    \includegraphics[width=0.8\linewidth]{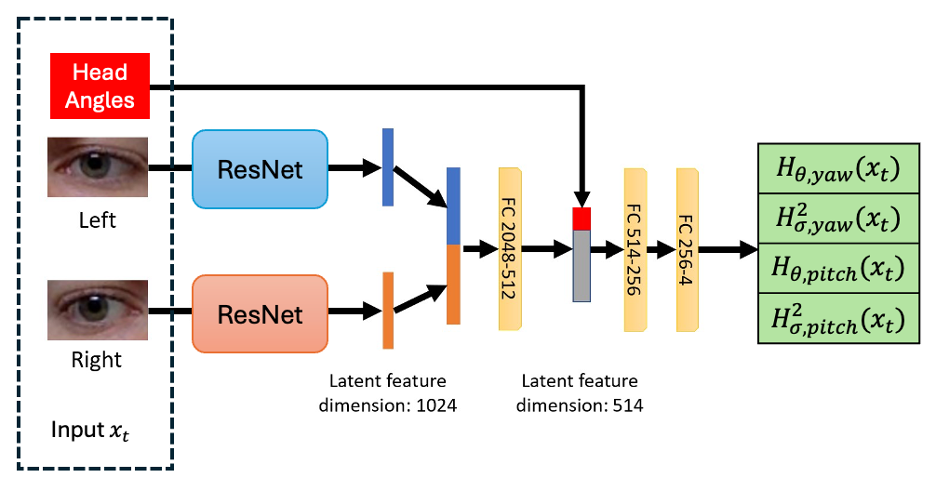}
    \caption{Uncertainty-aware gaze tracking model. It takes eye patches and head angles as input and outputs normal distributions for gaze tracking, with the mean capturing the gaze angle and the variance capturing uncertainty.}
    \label{fig:fig_4}
\end{figure}

The uncertainty-aware gaze estimator used in this study is a heteroskedastic regression model~\cite{Nix1994} that outputs probabilistic descriptions of gaze angles. This unmodified approach, widely adopted in prior works~\cite{Chen2019, Her2023}, provides uncalibrated uncertainty estimations that serve as the baseline. The model assumes gaze angles follow Gaussian distributions and outputs their means and variances, the variances represent model uncertainty. The training process builds a forecaster $H$ that takes inputs $x_t$ composed of left eye patch, right eye patch, and head angle readings and outputs two Gaussian distributions, described by mean values $H_{\theta,yaw}\left(x_t\right)$,$\ H_{\theta,pitch}(x_t)$ and variance values $H_{\sigma,yaw\ }^2\left(x_t\right)$,  $H_{\sigma,pitch\ }^2\left(x_t\right)$, respectively, as shown in Figure \ref{fig:fig_4}. For simplicity, we denote the estimated means and variances as $\hat{\theta}_t$ and $\hat{\sigma}_t^2$, respectively, each with pitch and yaw components, and the true gaze angles as ${\theta}_t$. Two network variants, ResNet-18 and ResNet-50, were implemented to examine calibration effectiveness across different model capacities.

The model uses a negative log-likelihood (NLL) loss (Equation~\ref{eqn:eqn6}) based on a heteroskedastic Gaussian distribution, where $l_n$ denotes the mean prediction error. For training stability, we use a smooth L1 loss (Equation~\ref{eqn:eqn7}) to measure this error. The NLL formulation enables training without variance labels by assuming a normal distribution. Although the true distribution may deviate from normality, this simple parametric assumption allows efficient probability computation and avoids the cost of MC dropout~\cite{gal2016dropout} or the complexity of Bayesian neural networks~\cite{NIPS2011_7eb3c8be, Blundell2015}.

\begin{equation}
\label{eqn:eqn6}
    NLL_t=\frac{1}{2}{ln}\left({\hat{\sigma}}_t^2\right)+\frac{l_{n,t}}{2{\hat{\sigma}}_t^2}
\end{equation}

\begin{equation}
\label{eqn:eqn7}
    l_{n,t}=\left\{\begin{array}{cc}
0.5(\hat\theta_t-\theta_t)^2, & \mathrm{ for }|\hat\theta_t-\theta_t|<1 \\
|\hat\theta_t-\theta_t|-0.5, & \mathrm{ otherwise}
\end{array}\right.
\end{equation}

\subsection{Evaluation Cases}
The evaluation considers four domain shift settings, formed by two cases, within-dataset cross-subject and cross-dataset, each applied to two datasets. In the within-dataset setting, all data splits—training, validation, calibration, and testing—come from the same dataset. In the cross-dataset setting, the model is trained and validated on one dataset, while calibration and testing use another. Calibration and testing remain subject-specific. Cross-dataset evaluation naturally introduces stronger domain shifts, whereas RTGene’s uniform noise yields milder shifts and MPIIGaze exhibits sharper anatomical detail. Performance is averaged across subjects in each fold, and each combination of dataset, evaluation setting, and CNN architecture is assessed using 3-fold cross-validation.

\section{Results and Discussion}

\subsection{Performance Improvements from Calibration}
\subsubsection{Uncertainty Estimation Accuracy Improvement}

\begin{figure}
    \centering
    \includegraphics[width=0.95\linewidth]{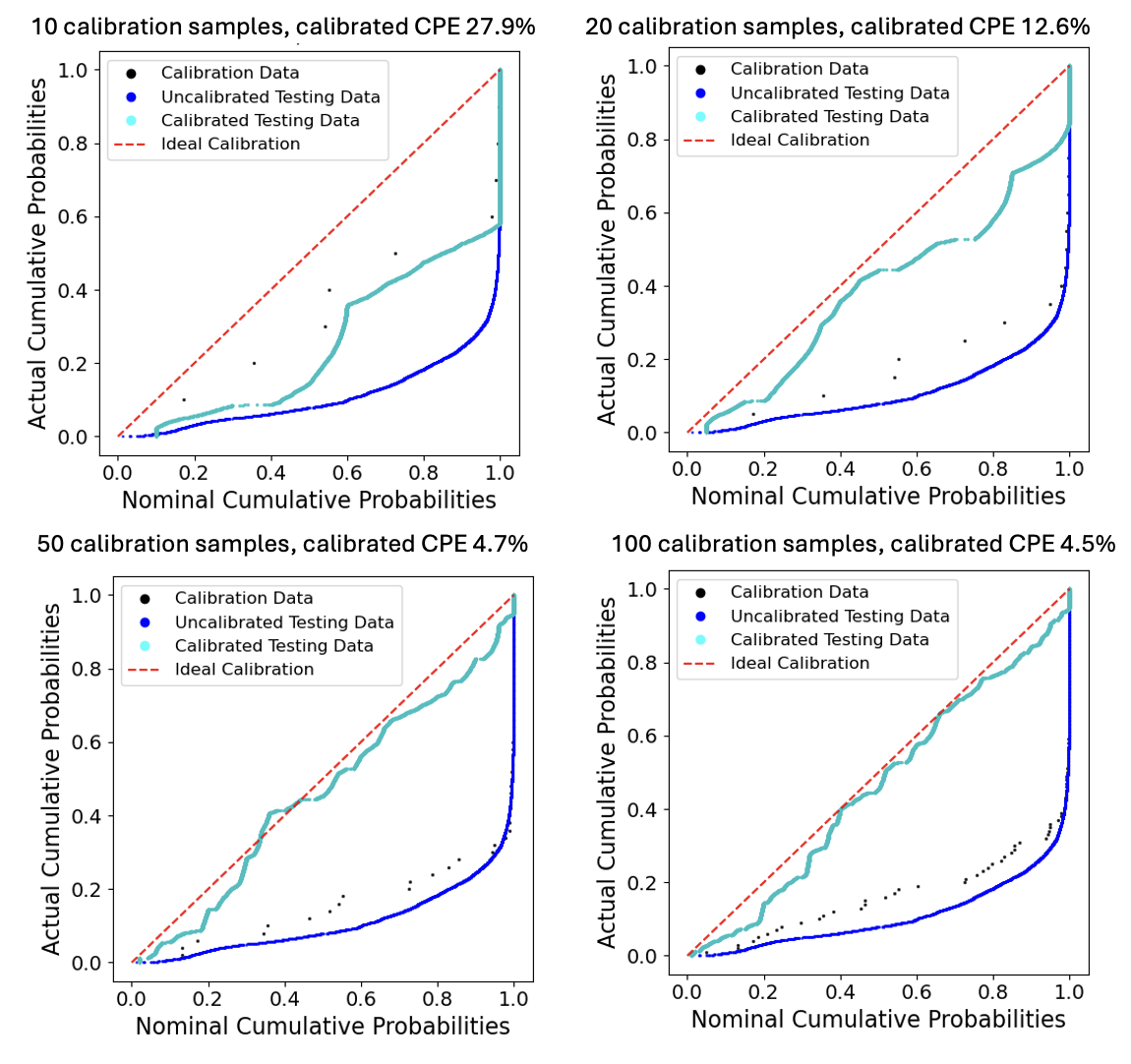}
    \caption{Influence of calibration sample sizes. More samples improve calibration. Performance increase plateaus around 50 samples. Uncalibrated model has CPE of 40.1\%. More results are included in supplementary video.}
    \label{fig:influence of calibration samples}
\end{figure}

\begin{figure}
    \centering
    \includegraphics[width=1.0\linewidth]{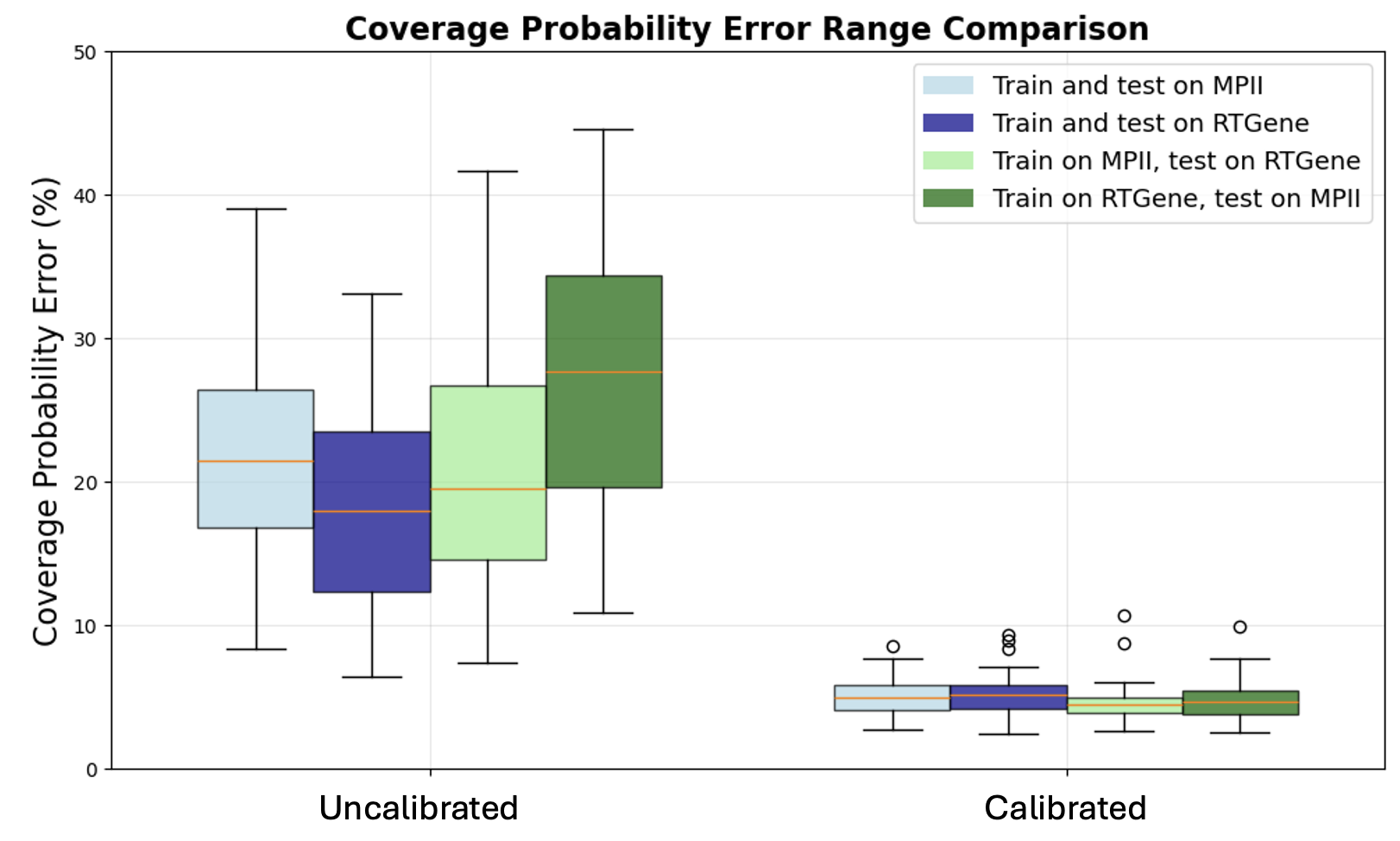}
    \caption{Distribution of CPEs. Orange lines mark medians, colored boxes show the interquartile range (25th–75th percentile), whiskers indicate the non-outlier bounds, and circles denote outliers. The narrower and lower distributions of calibrated models reflect more consistent and accurate uncertainty estimates, illustrating the significant and consistent improvement brought by the calibration method.}
    \label{fig:CPE distribution variation}
\end{figure}

Calibration effectively improves the accuracy of uncertainty estimation, as shown by the closer-to-ideal cyan curve in Figure~\ref{fig:influence of calibration samples}. This figure illustrates the effect of calibration sample count on model accuracy, using results from a test fold in the cross-subject configuration of the RTGene dataset. The uncertainty model was calibrated using 10, 20, 50, and 100 samples. More calibration samples enable a more accurate modeling of the miscalibration caused by domain shifts, leading to a more effective correction regressor $R$. Performance gains are most pronounced when the calibration set is small (around 10–20 samples), while gains plateau near 50 samples, where predictions approach the ideal case. Based on this observation, summary results presented in Tables~\ref{tab:table_2} and~\ref{tab:table_CI}, and Figure~\ref{fig:CPE distribution variation} use 50 calibration samples.

With calibration, both the accuracy and consistency of uncertainty estimates improve substantially across all four domain shift settings, as evidenced by the uniformly low CPE values in Figure~\ref{fig:CPE distribution variation}. The average CPE drops from a wide 8–45\% range to a stable level around 5\%, demonstrating robustness to varying degrees of domain shift. Average CPE changes across different domain shift settings are summarized in the corresponding columns in Table~\ref{tab:table_2}. All calibrated uncertainty models achieved over 70\% improvement, and these gains are statistically significant according to the Mann–Whitney U test.

\setlength{\tabcolsep}{2pt}
\renewcommand{\arraystretch}{0.9}
\begin{table*}
\centering
\caption{Comparison of Average CPE, Average EUC, and Average Angular Error for Evaluating Calibration}
\begin{tabular}{|c|c|c|c|c|c|c|c|c|c|c|}
\hline
\multicolumn{5}{|c|}{\diagbox{}{\hspace{3.1cm}}} & \multicolumn{2}{c|}{\makecell{Avg. CPE\\(percent)}} & \multicolumn{2}{c|}{ \makecell{Avg. EUC \\ (min:-1, max: 1)}} &\multicolumn{2}{c|}{\makecell{Avg. Angular Error\\(degree)}} \\ 
\hline
\makecell{Test\\Case} & \makecell{Test\\Type}&\makecell{Training\\ Dataset} &\makecell{Testing\\ Dataset} &Backbone& Uncal. & Calib. & Uncal. & Calib. &Uncal. & Calib.\\ 
\hline
1&\multirow{4}{*}{\makecell{Within\\Dataset, \\ Cross \\Subject}}&MPII & MPII & ResNet18 & 23.17 &  5.18$^{\dagger}$ $(\downarrow 78\%)$ & 0.14 & 0.26  & 5.77 & 5.09 $(\downarrow 12\%)$ \\
2& &RTGene & RTGene & ResNet18 & 19.60 &  5.26$^{\dagger}$ $(\downarrow 73\%)$ & 0.14 & 0.24  & 12.36 & 10.55 $(\downarrow 15\%)$ \\
3& &MPII & MPII & ResNet50 & 21.10 &  5.10$^{\dagger}$ $(\downarrow 76\%)$ & 0.05 & 0.18  & 6.04 & 5.53 $(\downarrow 8\%)$ \\
4& &RTGene & RTGene & ResNet50 & 17.50 &  5.21$^{\dagger}$ $(\downarrow 70\%)$ & 0.13 & 0.22  & 11.04 & 10.24 $(\downarrow 7\%)$ \\
\hline
5&\multirow{4}{*}{\makecell{Cross\\Dataset}}&MPII & RTGene & ResNet18 & 20.60 &  4.75$^{\dagger}$ $(\downarrow 77\%)$ & 0.04 & 0.09  & 13.71 & 10.12 $(\downarrow 26\%)$ \\
6& &RTGene & MPII & ResNet18 & 27.21 &  4.84$^{\dagger}$ $(\downarrow 82\%)$ & 0.15 & 0.13  & 18.46 & 14.50 $(\downarrow 21\%)$ \\
7 &&MPII & RTGene & ResNet50 & 20.10 &  4.63$^{\dagger}$ $(\downarrow 77\%)$ & 0.05 & 0.10  & 13.89 & 9.50 $(\downarrow 32\%)$ \\
8& &RTGene & MPII & ResNet50 & 26.36 &  4.79$^{\dagger}$ $(\downarrow 82\%)$ & -0.04 & 0.14  & 13.76 & 14.24 ($\uparrow 3\%$) \\
\hline
\end{tabular}
\label{tab:table_2}
\parbox{\linewidth}{\small Note: CPE values near zero reflect well-calibrated uncertainty models. EUC of 1 reflects perfect angular-error-uncertainty correlation, which was mistaken as perfect uncertainty model in~\cite{Kellnhofer2019, Her2023}. $^{\dagger}$ indicates a statistically significant difference between calibrated and uncalibrated results under Mann-Whitney U test ($p < 0.05$). p-values for EUC and angular errors are both 1. CPE from uncalibrated models serves as the baseline as similar uncalibrated uncertainty estimation models are widely used in gaze-tracking works~\cite{Her2023, Chen2019}. }
\end{table*}

\begin{table}
\centering
\captionsetup{justification=centering}
\caption{Comparison of Coverage Probability under Nominal 95\% Confidence Interval}
\begin{tabular}{|c|c|c|c|}
\hline
{\diagbox{}{\hspace{0.5cm}}} & \multicolumn{3}{c|}{Coverage Probability (percent)*} \\
\hline
Test Case& Quantile& Uncalibrated & Calibrated \\
\hline
1 & 40.5 & 41.1 & \textbf{88.0}\\
2 &51.6 & 51.4 & \textbf{86.5} \\
3 & 38.1 & 59.0 & \textbf{88.3}  \\
4 & 50.3 & 50.6 & \textbf{88.8} \\
\hline

5 & 34.3 & 47.8 & \textbf{86.7}\\
6 & 53.2 & 67.3 & \textbf{88.7} \\
7 & 34.0 & 48.5 & \textbf{89.0}  \\
8 & 16.4 & 46.2 & \textbf{88.6} \\
\hline
\end{tabular}
\label{tab:table_CI}
\parbox{1.0\linewidth}{\small* Coverage probability closer to 95\% or 0.95 is better. Quantile regression and uncalibrated models are the baselines.}
\end{table}



\subsubsection{Angular Error Improvement}
Although calibration primarily targets improving uncertainty estimation accuracy, the average angular error in Table~\ref{tab:table_2} reveals a secondary benefit of reducing angular prediction error. Calibration allows the predicted probability distribution to better approximate the true gaze-angle distribution rather than being constrained by assumed parametric forms, making median estimates more representative of actual angles. Most testing cases show a 7\%–32\% reduction in angular error, with only one case showing a slight 3\% increase. While less statistically significant than the CPE gains, the consistent error reduction across most cases still demonstrates that calibration reduces angular errors. Because angular error is not the main focus of this work, we do not analyze how calibration sample count affects it.

\subsection{CPE vs EUC for Uncertainty Assessment}

\begin{figure}
    \centering
    \includegraphics[width=1.0\linewidth]{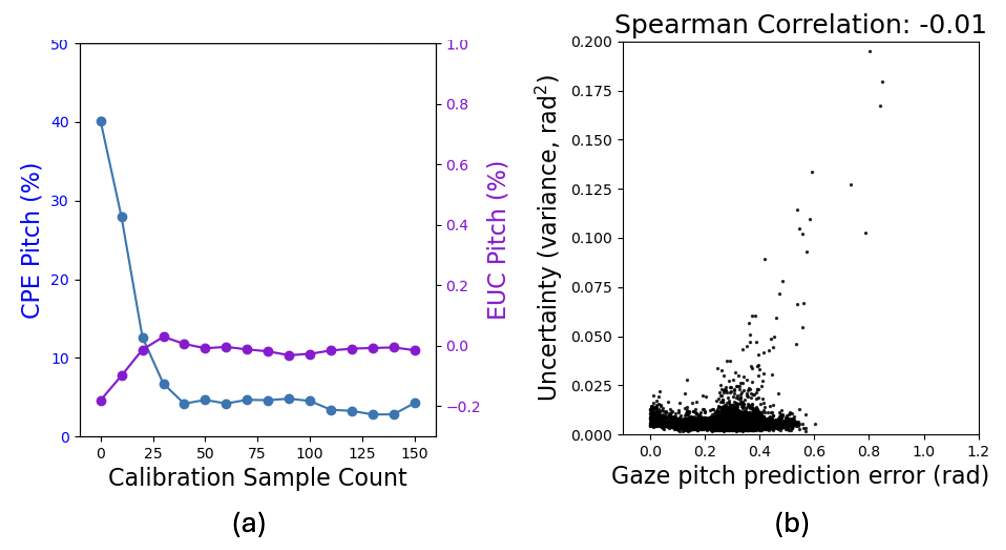}
    \caption{(a) Comparison of assessment effectiveness between CPE and EUC. CPE (blue curve) accurately reflecting model improvements shown in Figure~\ref{fig:influence of calibration samples}, assigning highly calibrated models near-perfect scores of 0. EUC (purple curve), however, fails to do so, misleadingly suggesting ineffective uncertainty modeling as its correlation strengths are near zero (1 is best for EUC). (b) Scatter plot showing prediction errors almost have no correlation with uncertainty with a near-zero EUC, indicating that angular errors are not the source of uncertainty, an observation supported by prior uncertainty modeling studies~\cite{kendall2017uncertainties}.}
    \label{fig:EUC vs CPE}
\end{figure}

A faithful assessment metric must capture the improvements shown in Figure~\ref{fig:influence of calibration samples}. The proposed CPE metric effectively reflects the performance gains achieved with increasing calibration samples, as illustrated by the blue curve in Figure~\ref{fig:EUC vs CPE}(a). The CPE value drops from around 40\% in the uncalibrated model to near zero for a well-calibrated model, accurately reflects the improvement. In contrast, EUC fails (purple line) to provide a reliable performance assessment. A near-zero correlation strength indicates that uncertainty is independent of angular prediction error, an observation that is consistent with existing uncertainty theories~\cite{kendall2017uncertainties}. However, according to prior uncertainty-aware studies~\cite{Kellnhofer2019, Her2023} based on EUC, it would erroneously suggest poor uncertainty modeling. This contradiction arises from the spurious correlations embedded in EUC. 

To visualize the underlying problem in EUC, a scatter plot (Figure~\ref{fig:EUC vs CPE}(b)) is created to visualize the correlation strength obtained from the near-ideal model, calibrated with 150 samples and achieving a CPE of about 5\%. Uncertainty exhibits little correlation with prediction error. Average EUC values are reported in the corresponding columns in Table~\ref{tab:table_2}. Across all test cases, EUC consistently produces conclusions that contradict the CPE evaluations. EUC values below 0.3 indicate weak or very weak correlation strength, falsely suggesting poor uncertainty models. In contrast, low CPE values indicate that the predicted probabilistic distributions closely match the true data, reflecting good model performance. Therefore, CPE provides a more faithful and reliable assessment than EUC.

\section{Case Study} \label{Case study}
To illustrate the practical benefits and include other uncertainty estimation approaches that are not compatible with CPE, we conducted a case study estimating the 95\% CI, a standard level used in mission-critical applications. The additional model included here is quantile regression~\cite{Kellnhofer2019}, because it does not produce a full distribution prediction, quantile regression is not suited for CPE. A well-calibrated model should produce CIs that cover the ground truth gaze angle labels with a 95\% probability. The coverage probabilities of estimated 95\% CI are summarized in Table~\ref{tab:table_CI}. CI estimations from both baseline models are significantly overconfident as the true coverage probabilities are often less than half of the expected value. CIs from the calibrated models are much closer to the expected 95\% value, providing more reliable estimation. 

\begin{figure}
    \centering
    \includegraphics[width=1.0\linewidth]{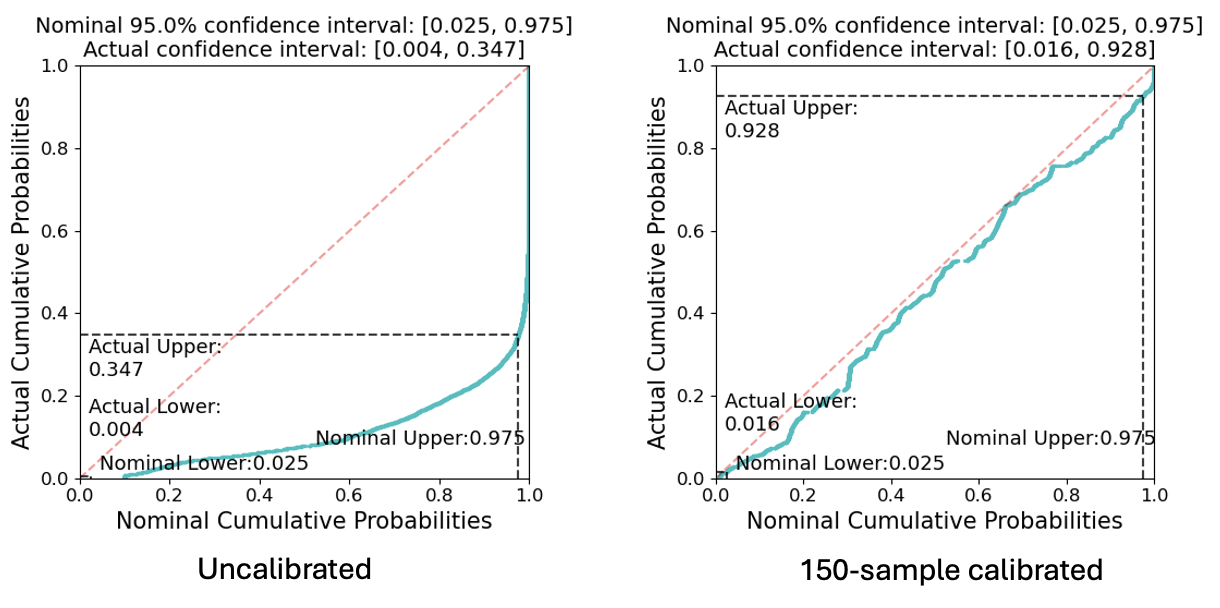}
    \caption{Comparison of 95\% confidence intervals (CIs) from uncalibrated and calibrated uncertainty models. The calibrated model yields coverage probabilities closer to the nominal 95\% due to its better-aligned probabilistic distribution.}
    \label{fig:Case study}
\end{figure}

We further visualize the sources of high CI errors in uncalibrated models in Figure~\ref{fig:Case study}. The probabilistic distributions from uncalibrated models deviate substantially from the ideal line, leading to quantile predictions that fail to meet expected coverage levels. After calibration, the distributions align more closely with the ideal, allowing the predicted quantiles to better reflect the true distribution.

\section{Conclusion}
In this study, we tackle inaccurate uncertainty estimation under domain shifts in appearance-based gaze tracking using a post-hoc calibration method that requires no modification to the original model. We also present a proper evaluation metric inspired by Brier score, called CPE, to provide a more faithful assessment of uncertainty estimation compared to the existing EUC metric. Experiments conducted on two widely used gaze datasets, MPIIGaze and RTGene, demonstrate the improvements achieved through calibration and the increased effectiveness of CPE in evaluating uncertainty estimation performance.


{
    \small
    \bibliographystyle{ieeenat_fullname}
    \bibliography{aaai2026}
}

\end{document}